%% file: main.tex
\documentclass[twoside]{article}
\usepackage{aistats2018}
\usepackage{times}
\usepackage{graphicx} 
\usepackage{subfigure} 

\usepackage[numbers]{natbib}

\usepackage{algorithm}
\usepackage{algorithmic}

\usepackage{hyperref}

\usepackage{multirow}
\usepackage{hhline}

\usepackage{amsmath,amssymb,amsthm}

\usepackage{bm} 

\def\P{\mathcal{P}}
\def\w{\vct{w}}
\def\z{\vct{z}}

\def\x{{\vct{x}}}

\def\blambda{{\boldsymbol \lambda}}

\def\P{\mathbb{P}}
\def\H{\mathcal{H}}

\def\W{\mathcal{W}}
\def\L{\mathcal{L}}
\def\G{\mathcal{G}}

\newcommand{\eat}[1]{{}}
\newtheorem{theorem}{Theorem}

\usepackage{array}

\newcolumntype{M}[1]{>{\centering\arraybackslash}m{#1}}
\newcolumntype{N}{@{}m{0pt}@{}}

\input{math_definition.tex}

\usepackage{tikz}
\usetikzlibrary{calc,arrows,positioning,shapes}

\begin{document}

%

%

\twocolumn[

\aistatstitle{Infinite-Label Learning with Semantic Output Codes}

\aistatsauthor{ Yang Zhang$^{1}$ \And Rupam Acharyya$^{2}$ \And  Ji Liu$^{3}$ \And Boqing Gong$^{4}$ }

\aistatsaddress{   Center for Research in Computer Vision\\
  University of Central Florida\\
  Orlando, FL 32816 \\
  \texttt{yangzhang@knigths.ucf.edu}$^{1}$ \\ \texttt{bgong@crcv.ucf.edu}$^{4}$\And Department of Computer Science\\
  University of Rochester\\
  Rochester, NY 14627\\
  \texttt{racharyy$^{2}$,jliu$^{3}$@cs.rochester.edu}   } ]

\begin{abstract} 
\input{abs}
\end{abstract} 

\input{intro}

\input{approach}

\input{theorem}
\input{exp}

\input{conclusion}

\end{document}


\twocolumn[
\icmltitle{Supplementary Materials for \\ Zero-Shot Multi-Label Learning with Vectorial Labels}



\icmlsetsymbol{equal}{*}

\begin{icmlauthorlist}
\icmlauthor{Aeiau Zzzz}{equal,to}
\icmlauthor{Bauiu C.~Yyyy}{equal,to,goo}
\icmlauthor{Cieua Vvvvv}{goo}
\icmlauthor{Iaesut Saoeu}{ed}
\icmlauthor{Fiuea Rrrr}{to}
\icmlauthor{Tateu H.~Yasehe}{ed,to,goo} 
\icmlauthor{Aaoeu Iasoh}{goo}
\icmlauthor{Buiui Eueu}{ed}
\icmlauthor{Aeuia Zzzz}{ed}
\icmlauthor{Bieea C.~Yyyy}{to,goo}
\icmlauthor{Teoau Xxxx}{ed}
\icmlauthor{Eee Pppp}{ed}
\end{icmlauthorlist}

\icmlaffiliation{to}{University of Torontoland, Torontoland, Canada}
\icmlaffiliation{goo}{Googol ShallowMind, New London, Michigan, USA}
\icmlaffiliation{ed}{University of Edenborrow, Edenborrow, United Kingdom}

\icmlcorrespondingauthor{Cieua Vvvvv}{c.vvvvv@googol.com}
\icmlcorrespondingauthor{Eee Pppp}{ep@eden.co.uk}

\icmlkeywords{zero-shot, multi-label}

\vskip 0.3in
]



\printAffiliationsAndNotice{} 

\input{theorem-supp}

\eat{
\bibliography{infinite}
\bibliographystyle{icml2017}
}

%% file: math_definition.tex
\usepackage{amsmath,amssymb,amsopn}
\usepackage{bm} 

\newcommand{\vct}[1]{\boldsymbol{#1}} 
\newcommand{\mat}[1]{\boldsymbol{#1}} 

\newcommand{\vmu}{\vct{\mu}}

\newcommand{\vlambda}{\vct{\lambda}}

\newcommand{\vf}{\vct{f}}

\newcommand{\vw}{\vct{w}}
\newcommand{\vx}{\vct{x}}
\newcommand{\vy}{\vct{y}}

\newcommand{\mU}{\mat{U}}

\newcommand{\field}[1]{\mathbb{#1}}
\newcommand{\R}{\field{R}} 

\newcommand{\Map}[1]{\mathcal{#1}}

\newcommand{\calF}{\Map{F}}

\newcommand{\calH}{\Map{H}}

\newcommand{\calL}{\Map{L}}

\newcommand{\calN}{\Map{N}}

\newcommand{\calS}{\Map{S}}

\newcommand{\calU}{\Map{U}}

\newcommand{\cst}[1]{\mathsf{#1}}

\newcommand{\cK}{\cst{K}}
\newcommand{\cL}{\cst{L}}
\newcommand{\cM}{\cst{M}}

\newcommand{\inner}[2]{\left\langle#1, #2\right\rangle}
\newcommand{\norm}[1]{\left\|#1\right\|}

\newcommand{\E}{\field{E}} 



%% file: abs.tex

We formalize a new statistical machine learning paradigm, called {\em infinite-label learning}, to annotate a data point with more than one relevant labels from a candidate set which pools both the finite labels observed at training and a potentially infinite number of previously unseen labels. The infinite-label learning fundamentally expands the scope of conventional multi-label learning and better meets the practical requirements in various real-world applications, such as image tagging, ads-query association, and article categorization. However, {how can we learn a labeling function that is capable of assigning to a data point the labels omitted from the training set?} To answer the question, we seek some clues from the recent works on zero-shot learning, where the key is to represent a class/label by a vector of semantic codes, as opposed to treating them as atomic labels. We validate the infinite-label learning by a PAC  bound in theory and some empirical studies on both synthetic and real data.

\eat{
Zero-shot learning has rapidly gained popularity in the recent years. It is appealing as it may bring machine learning closer to one of the remarkable human learning capabilities: the generalization from past experience is not only to new data but also to new tasks. Despite the extensive existing work on zero-shot learning for multi-class classifications, it is rarely studied for multi-label classification. We contend that it is actually indispensable for multi-label learning algorithms to handle previously unseen labels. It is often more costly to collect clean training data for multi-label learning than for classification problems. In this paper, we provide a preliminary and formal treatment to the \textbf{zero-shot multi-label learning} problem. We formalize it as a {general} machine learning framework and show its feasibility both theoretically under some mild assumptions and empirically on synthetic and real data. Our results indicate that zero-shot multi-label learning is hard and yet not impossible to solve given appropriate vectorial representations of the labels. We expect the progress on this problem can significantly advance novel approaches that are data-efficient from the training perspective, and are capable of dealing with previously unseen labels at the test phase.
} 

%% file: intro.tex

\section{Introduction}
Recent years have witnessed a great surge of works on zero-shot learning~\cite{larochelle2008zero,palatucci2009zero,frome2013devise,socher2013zero,lampert2014attribute,lei2015predicting,romera2015embarrassingly,chao2016empirical}. While  conventional supervised learning methods (for classification problems) learn classifiers only for the classes that are supported by data examples in the training set, the zero-shot learning also strives to construct the classifiers of novel classes for which there is no training data at all. This is achieved by using class/label descriptors, often in the form of some vectors of numeric or symbolic values, in contrast to the atomic classes used in many other machine learning settings. 

Zero-shot learning is appealing as it may bring machine learning closer to one of the remarkable human learning capabilities: generalizing learned concepts from old tasks to not only new data but also new (and yet similar) tasks~\cite{moses1996generalization,hancock2000recognition}. There are also many real-world applications for zero-shot learning. It facilitates cold-start~\cite{schein2002methods} when the data of some classes are not available yet. In computer vision, the number of available images per object follows a long-tail distribution, and zero-shot learning becomes key to handling the rare classes. We note that the success of zero-shot learning largely depends on how much knowledge the label descriptors encode for relating the novel classes with the seen ones at training, as well as how the knowledge is involved in the decision making process, e.g., for classification. 

Despite the extensive existing works on zero-shot learning for multi-class classifications, it is rarely studied for multi-label classification~\cite{tsoumakas2009mining,zhang2014review,bhatia2015sparse} except by~\cite{nam2015predicting,Zhang_2016_CVPR}. We contend that it is actually indispensable for multi-label learning algorithms to handle previously unseen labels. There are always new topics arising from the news (and research, education, etc.) articles over time. Creative hashtags could become popular on social networks over a night. There are about 53M tags on Flickr and many of them are associated with none or very few images. Furthermore, it is often laborious and costly to collect clean training data for the multi-label learning problems especially when there are many labels~\cite{bi2013efficient,bhatia2015sparse}. To this end, we contend that zero-shot multi-label learning, which we call {\em infinite-label learning} to make it concise and  interesting, is arguably a much more pressing need than the zero-shot multi-class classification.

To this end, we provide a  formal treatment to the {infinite-label learning} problem in this paper. To the best of our knowledge, the two existing works~\cite{nam2015predicting,Zhang_2016_CVPR} on this problem only study {specific} application scenarios. Zhang et al.\ analyze the rankability of word embeddings~\cite{mikolov2013distributed,pennington2014glove} and then use them to assign both seen and unseen tags to images~\cite{Zhang_2016_CVPR}. Nam et al.\ propose a regularization technique for the WSABIE algorithm~\cite{weston2011wsabie} to account for the label hierarchies~\cite{nam2015predicting}. In sharp contrast, we formalize the infinite-label learning as a {general} machine learning framework and show its feasibility both theoretically under some mild assumptions and empirically on synthetic and real data. Our results indicate that the infinite-label learning is hard and yet not impossible to solve, given  appropriate semantic codes of the labels. 

The semantic codes of the labels can be derived from a large knowledge base in addition to the training set. Thanks to the studies in linguistics, words in speech recognition are represented by combinations of phonemes~\cite{waibel1989modular}. In computer vision, an object class is defined by a set of visual attributes~\cite{lampert2014attribute,farhadi2009describing}. More recently, the distributed representations of English words~\cite{mikolov2013distributed,pennington2014glove} have found their applications in a variety of tasks. In biology, the labels are naturally represented by some structural or vectorial codes (e.g.,  protein structures~\cite{stock2016efficient}).

 
We organize the remaining of the paper as follows. We discuss some related work and machine learning settings next. After that, Section~\ref{sProblem} formally states the problem of infinite-label learning, followed by some plausible modeling assumptions and algorithmic solutions. We then present a PAC-learning bound and some empirical results for this new machine learning framework in Sections~\ref{sPAC} and \ref{sEmpirical}, respectively. Section~\ref{sConclusion} concludes the paper.

\section{Related work}

We note that the semantic codes of labels have actually been used in various machine learning settings even before the zero-shot learning becomes popular~\cite{larochelle2008zero,palatucci2009zero}. Bonilla et al.\  use the vectorial representations of tasks to enrich the kernel function and parameterize a gating function~\cite{bonilla2007kernel}. Similarly, Bakker and Heskes input the task vectors to a neural network for task clustering and gating~\cite{bakker2003task}. Wimalawarne et al.\ address multi-task learning when the tasks are each indexed by a pair of indices~\cite{wimalawarne2014multitask}. A large-margin algorithm is introduced in~\cite{szedmak2006learning} to transform the classification problems to multivariate regression by considering the semantic codes of the classes as the target. It is worth pointing out that this line of works, unlike zero-shot learning or the infinite-label learning studied in this paper, lack the exploration of the learned models' generalization capabilities to novel labels or tasks that are unseen in training.

Learning to rank~\cite{stock2014identification,pahikkala2013efficient} and the cold-start problem in recommendation system~\cite{adams2010incorporating,zhou2012kernelized,fang2011matrix} explicitly attempt to tackle unseen labels (e.g., queries and items, respectively) thanks to the label descriptions that are often vectorial. Zero-shot learning shares the same spirit~\cite{larochelle2008zero,palatucci2009zero,lampert2014attribute,frome2013devise,socher2013zero,romera2015embarrassingly,lei2015predicting,chao2016empirical,akata2015evaluation,fu_transductive_2014} and we particularly study the  infinite-label learning in this paper.

Additionally, the semantic codes of labels or tasks also seamlessly unify a variety of machine learning settings ~\cite{yang2014unified,stock2016efficient}. The infinite-label learning studied in this paper can be regarded as a special case of the pairwise learning; in particular, a special case of the setting D in~\cite{stock2016efficient}.  Nonetheless, we think the infinite-label learning is worth studying as a standalone machine learning framework given its huge application potentials. Besides, the PAC-learning analysis presented in this paper sheds light on the learnability of the pairwise learning~\cite{stock2016efficient}.

The recent works of label embedding~\cite{hsu2009multi,weston2011wsabie,zhang2011multi,bhatia2015sparse} and landmark label sampling~\cite{balasubramanian2012landmark,bi2013efficient} in multi-label learning are also related to our work. The main distinction is that their label representations are learned from the training set and are not applicable to unseen labels. 


\eat{

We formalize a new machine learning paradigm, {infinite}-label learning, where the goal is to learn a labeling function to annotate a data point with all relevant labels from a candidate set, which pools both the finite labels observed at training and a potentially infinite number of previously unseen labels. Infinite-labeling learning is motivated by a plethora of real-world applications. As below, we give a few examples to facilitate our further discussion. 

{\bf Image tagging.} More than 50\% of the Flickr images have no text tags and are never retrieved for text queries. As a result, automatically suggesting relevant tags/labels for the images is a pressing need and yet a daunting challenge; there are more than 53M tags on Flickr. Moreover, tons of new tags emerge over time. 

{\bf Ads-query association.} Search engines allow advertisers to bid on user queries to deliver ads to targeted audience. It is both commercially intriguing and scientifically interesting to automatically associate the virtually infinite number of user queries (labels) to the ads (data points). 

{\bf Text categorization.} The \textsc{wikipedia} pages are grouped by more than 200K categories. A text might be about any of science, politics, finance, and/or other topics. Infinite-label learning can save the writers from the tedious process of manually classifying the future \textsc{wikipedia}  pages, even if new categories are added to \textsc{wikipedia}.

The above problems are actually often used to benchmark different (extreme) multi-label learning algorithms. However, we contend that the conventional multi-label learning tackles the problems on the surface, and fails to consider any new labels that appear in the test stage and were omitted from the training stage. In a sharp contrast, \emph{infinite}-label learning fundamentally expands the scope of \emph{multi}-label learning; it is expected to explicitly handle novel labels that show up after the labeling function has been learned.

\emph{How can we learn a labeling function that is capable of assigning to a data point the labels  omitted from the training set?} To answer the question, we seek some clues from the recent work on zero-shot learning~\cite{palatucci2009zero,socher2013zero,frome2013devise,romera2015embarrassingly}, where the key is to represent a class by a vector of semantic codes~\cite{palatucci2009zero}, as opposed to treating them as atomic labels. We can thus learn a labeling function to take as input a data point and a label's semantic codes, and output whether or not the encoded label is relevant to the data. Both seen and unseen labels are encoded by the same mechanism and usually by the same knowledge base. As a result, the labeling function learned from a finite training set is able to extrapolate to the labels unseen during the training. To verify this intuition, we provide a PAC  bound (Section~\ref{sPAC}) and some empirical studies (Section~\ref{sEmpirical}) on both synthetic and real data.

The semantic codes of the class labels are often derived from a large knowledge base in addition to the training set. Thanks to the studies in linguistics, words in speech recognition are represented by combinations of phonemes~\cite{waibel1989modular}. In visual object recognition, a class is described by a set of visual attributes~\cite{lampert2014attribute,farhadi2009describing}. More recently, the distributed representations of English words~\cite{mikolov2013distributed,pennington2014glove} have found their applications in a variety of tasks. 
}

%% file: approach.tex

\section{Infinite-label learning} \label{sProblem}
The goal of infinite-label learning is to learn a decision function to annotate a data point with all relevant labels from a candidate set which pools both the finite labels observed at training and a potentially infinite number of previously unseen labels. In this section, we formally state the problem and then lay out a modeling assumption about the data generation process. We also provide some potential algorithmic solutions to the infinite-label learning. 

\begin{table*}
\centering
\small
\caption{{Multi}-label learning \textit{vs.}\ infinite-label learning.} \label{tInfLabel}
\vspace{3pt}
\begin{tabular}{|l|M{6cm}|M{6cm}|@{}m{0pt}@{}}
\hline
&  Multi-label learning & infinite-label learning (ZSML)  &\\ [10pt]
\hline
Data generation &
\begin{tikzpicture}[
    node distance=2em,
    thick,
    terminal/.style={draw,circle,inner sep=1pt,minimum size=1.3em},
    termell/.style={draw,shape=ellipse,inner sep=1pt,minimum height=1.4em},
    doubletrans/.style={->,bend left=20}
  ]
  \small
\node (s0) [terminal] at (0,0)		{$X$};
\node (s1) [termell,right=of s0] 	{$\mathcal{L}=(\blambda_1,\blambda_2,\cdots,\blambda_\cL)$};
\draw [->] (s0) to node[right] {$$} (s1);
 \end{tikzpicture} & \small
\begin{tikzpicture}[
    node distance=2em,
    thick,
    terminal/.style={draw,circle,inner sep=1pt,minimum size=1.3em},
    termell/.style={draw,shape=ellipse,inner sep=1pt,minimum height=1.4em},
    doubletrans/.style={->,bend left=20}
  ]
\node (s0) [terminal] at (0,0)		{$X$};
\node (s1) [terminal,right=of s0] 		{$Y$};
\node (s2) [terminal,right=of s1] 	{$\Lambda$};
\draw [->] (s0) to node[right] {$$} (s1);
\draw [->] (s2) to node[left] {$$} (s1);
 \end{tikzpicture} &\\ [12pt]
Training data & $\calS=\{(\vx_m,\vy_m)\}_{m=1}^\cM$ & $\calS, \; \calL=\{\vlambda_l\}_{l=1}^\cL$ &\\ [7pt]
Labeling function & $\vf(\vx)\subseteq\calL$ & $\vf(\vx)\subseteq\calL\cup\calU$, $\;\calU=\{\vlambda_l\}_{l>\cL}$ &\\ [7pt]
Example solution & $f_l(\vx)=\text{sgn}\inner{\vw_l}{\vx}, \; 1\le l \le\cL$  & $f_l(\vx)=f({\vlambda}_l;\vx) = \text{sgn}\inner{V\vx}{{\vlambda}_l}, \; l\ge 1$ &\\ [7pt]
 \hline
\end{tabular}
\vspace{-0pt}
\end{table*}

\subsection{Problem statement}
Suppose we have a set of vectorial labels $\calL=\{\vlambda_l\in\R^\cst{n}\}_{l=1}^\cL$ and a training sample $\calS=\{(\vx_m,\vy_m)\in\R^\cst{d}\times\{-1,+1\}^\cL\}_{m=1}^\cM$, where the annotation $y_{ml}=+1$ indicates that the $l$-th label, which is described by $\vlambda_l$, is relevant to the $m$-th data point $\vx_m$, and $y_{ml}=-1$ otherwise. For convenience, we refer to the label vectors $\calL=\{\vlambda_l\}_{l=1}^\cL$ as labels in the rest of the paper. Unlike conventional multi-label learning where one learns a decision function $\vf({\vx})\subseteq\calL$ to assign the seen labels to an input data point, the infinite-label learning enables the decision function to additionally tag the data point with previously unseen labels $\calU=\{\vlambda_l\}_{l>\cL}$, where $\calL\cap\calU=\emptyset$.  

Table~\ref{tInfLabel} conceptually compares the infinite-label learning with multi-label learning. The rows of training data and labeling functions summarize the above discussions, and the other rows are elaborated in the next two subsections, respectively.


\subsection{The non-i.i.d.\ training set} \label{sNonIID}
One of our objectives in this paper is to derive a PAC-learning bound for the infinite-label learning. To facilitate that, we carefully examine the data generation process in the infinite-label learning in this subsection. 

At the first glance, one might impose a distribution $P_{X{\Lambda}Y}=P_{X{\Lambda}}P_{Y|X\Lambda}$ over the data $\vx\sim X$, vectorial label $\vlambda\sim \Lambda$, and the indicator $y\sim Y|{\vx,\vlambda}$ about the relevance between the label and the data, and then assume the training sample ($\calS$ and $\calL$ jointly) is drawn \textbf{i.i.d.}\ from $P_{X{\Lambda}Y}$. Noting that $y$ implies nothing but a binary classification over the augmented data $(\vx,\vlambda)$, the existing PAC-learning results could be used to bound the generalization error of the infinite-label learning. 

The above reasoning is actually problematic because its corresponding generalization risk cannot be estimated empirically in real scenarios and applications. To see this point more clearly, we write out the generalization risk and its straightforward empirical estimate below, 
\begin{align}
R(h) &= \E_{\vx,\vlambda\sim P_{X\Lambda}}\E_{y\sim P_{Y|\vx,\vlambda}
}[h(\vx, \vlambda)\neq y] \\
\widehat{R}_\text{iid}(h) &=\frac{1}{\cM} \sum_{m=1}^\cM [h(\vx_m, \vlambda_m)\neq y_m]
\end{align}
where $h\in\calH$ is a hypothesis from a pre-fixed family $\calH$ and $[\cdot]$ is the 0-1 loss. We can see that, under the i.i.d.\ assumption, the data and label $(\vx_m,\vlambda_m)$ must be sampled simultaneously, i.e., they are exclusively coupled. 

However, the seen labels and data in the training set are often pairwise examined and are not exclusively coupled. For example, to build the NUS-WIDE dataset~\cite{chua2009nus}, the annotators are asked to give the relevance between all the 81 tags and each image. Such practice instead suggests an alternative empirical risk, 
\begin{align} \label{eEmpRisk}
\widehat{R}_\text{non-iid}(h) = \frac{1}{\cM\cL}\sum_{m=1}^\cM \sum_{l=1}^\cL [h(\vx_m, \vlambda_l)\neq y_{ml}]. \nonumber
\end{align}
Moreover, the decoupling of  data example $\vx_m$ and label $\vlambda_l$ in $\widehat{R}_\text{non-iid}(h)$ also implies that the training set in infinite-label learning (and multi-label learning) is virtually generated from the joint distribution $P_{X{\Lambda}Y}$ in the following non-i.i.d.\ manner. 
\vspace{-5pt}
\begin{itemize}   \setlength\itemsep{0.1em}
\item Sample $\cM$ times from $P_X$ and obtain $\{\vx_m\}_{m=1}^\cM$,
\item Sample $\cL$ times from $P_{\Lambda|\vx_1,\cdots,\vx_\cM}$ and obtain $\{\vlambda_l\}_{l=1}^{\cL}$,
\item Sample $y_{ml}$ from $P_{Y|\vx_m,\vlambda_l}$ for $m\in[\cM], l\in[\cL]$.
\end{itemize}

As a result, the training set is a \textbf{non-i.i.d.}\ sample drawn from $P_{X{\Lambda}Y}$. The existing PAC-learning bounds for binary classification cannot be directly applied here. Instead, we would like to bound the difference between the generalization risk $R(h)$ and the  non-i.i.d.\ empirical risk $\widehat{R}_\text{non-iid}(h)$. To overcome this disparity is accordingly another contribution of this paper presented in Section~\ref{sPAC}.


\paragraph{Remarks.} We note that the joint distribution $P_{X\Lambda{Y}}$ is in general not sufficient to specify $P_{\Lambda|\vx_1,\cdots,\vx_\cM}$ appeared in the above sampling procedure. To rectify this, we make an independent assumption detailed in Section~\ref{sPAC}.



\subsection{Views of the problem and algorithmic solutions}
Given the sampling procedure described above, to solve infinite-label learning essentially is essentially to approximate the conditional distribution $P_{Y|X\Lambda}$. One potential hypothesis set for $P_{Y|X\Lambda}$ takes the simple bilinear form $f(\vx,\vlambda)=\text{sgn}\inner{V\vx}{\vlambda}$. Following the discussion about zero-data learning~\cite{larochelle2008zero}, we may understand this decision function from at least two views. From the data point of view, it is  a decision function defined over the augmented input data, i.e., $(\vx,\vlambda)$. From the model point of view, it defines a hyperplane $V\vx$ that separates the labels $\vlambda\in\calL\cup\calU$ into two subsets no matter they are seen or unseen. One subset of the labels is relevant to the data instance $\vx$ and the other is not, thanks to that the hyperplane is indexed by the input data. To learn the model parameters $V$, the maximum margin regression~\cite{szedmak2006learning} may be readily applied.

We can also use a neural network $\textsc{nn}(\vx)$ to infer the hyperplanes in the label space, such that $f(\vx,\vlambda)=\text{sgn}\inner{\textsc{nn}(\vx)}{\vlambda}$. Additionally, the kernel methods reviewed in~\cite{stock2016efficient} for pairwise learning are natural choices for solving the infinite-label learning. In general, the kernel function gives rise to the following form of decision functions: $f(\vx,\vlambda)=\sum_{l=1}^\cL\alpha_l(\vx)k(\vlambda,\vlambda_l)$, where $\alpha_l(\cdot)$ takes as input a data point $\vx$ and outputs a scalar and $k(\cdot,\cdot)$ is a kernel function over the labels. 

In the following, we will mainly study the bilinear decision function $f(\vx,\vlambda)=\text{sgn}\inner{V\vx}{\vlambda}$ both theoretically (Section~\ref{sPAC}) and empirically (Section~\ref{sEmpirical}). 


%% file: theorem.tex

\section{A PAC-learning analysis} \label{sPAC}
In this section, we investigate the learnability of the infinite-label learning under the PAC-learning framework. Given the training set $\calS=\{(\vx_m,\vy_m)\in\R^\cst{d}\times\{-1,+1\}^\cL\}_{m=1}^\cM$ and the \emph{seen} vectorial labels $\calL=\{\vlambda_l\in\R^\cst{n}\}_{l=1}^\cL$, the theorem below sheds light on the numbers of data points and labels that are necessary to achieve a particular level of error of the generalization  to {\em not only test data examples but also previously unseen labels}. 

Our result depends on two mild assumptions.

\paragraph{Assumption I: $P_{\Lambda|X}=P_{\Lambda}$.} Recall the second step in the sampling procedure in Section~\ref{sNonIID}. The conditional distribution $P_{\Lambda|\vx_1,\cdots,\vx_\cM}$ is intractable in general from only the joint distribution $P_{X\Lambda}$. To rectify this, we introduce the first assumption here, that knowing the data does not alter the marginal distribution of the labels, i.e., $P_{\Lambda|X}=P_{\Lambda}$. This is reasonable for the infinite-label learning especially when the label descriptions come from a knowledge base that is distinct from the training set $\calS$ --- e.g., the word embeddings used in~\cite{Zhang_2016_CVPR} are learned without accessing the images, and the label hierarchies are given independent of the news articles in~\cite{nam2015predicting}. 

Due to \textbf{Assumption I}, we arrive at a simplified modeling about the infinite-label learning problem. Table~\ref{tInfLabel} draws the corresponding graphical model from which we can see that the distribution over the indicator variable $Y$ is unique from the graphical model of multi-label learning.  It is this change of modeling that enables the generalization to new labels feasible.




\paragraph{Assumption II:} The conditional distribution of the indicator variable $P_{Y|\vx,\vlambda}$ is controlled by a random variable through $y|\vx,\vlambda=\sigma y^\star(\vx,\vlambda)$  where $\sigma\in\{1,-1\}$ is a binary random variable and $y^\star(\vx,\vlambda)$ is the noiseless indication about the relevance of the label $\vlambda$ with the data $\vx$. 

This assumption partially grounds on that, in practice, the annotations of the training sample $\calS$ are often incomplete when the number of seen labels $\calL$ is large. Take the user-tagged data in social networks for instance. Some labels for which $\{y_{ml}=-1\}$ are actually relevant to the data; the indicators $\{y_{ml}\}$ are flipped to $-1$ (from the groundtruth $+1$) merely because the corresponding labels fail to capture the users' attentions and are thus missed by the users. 


\eat{
We consider the following distributions over the data in infinite-label learning, 
\begin{align}
&\vx\sim P_X,\quad \vlambda\sim P_\Lambda, \quad y(\vlambda;\vx)\sim P_{Y|X{\Lambda}}, \\
&\quad \text{ where } P_{X{Y}\Lambda}=P_X P_\Lambda P_{Y|X{\Lambda}},
\end{align} 
where $P_{X{Y}\Lambda}$ denotes the joint distribution of the data point $\vx$, label assignment indicators $\vy$, and the label $\vlambda$. In this paper, we consider a deterministic/noiseless label assignment function $y(\vlambda;\vx)\sim P_{Y|X{\Lambda}}$, and leave the more generic case to the future work.

Note that we explicitly treat the labels $\calL\cup\calU$ as an i.i.d.\ sample drawn from the marginal distribution $P_\Lambda$. Moreover, we make the following observation: any data point $\vx$ can be regarded as incurring a binary classification rule, denoted by $y(\cdot;\vx)$, over the label space, such that $y(\vlambda;\vx)=+1$ tells that the label $\vlambda\sim P_\Lambda$ is relevant to the data $\vx$ and $y(\vlambda;\vx)=-1$ means the opposite. This rule is only partially observed at training in the form of the data annotations $\vy_m$.

Immediately following the above modeling assumption, our learning objective is  to infer the (conditional) classification rule $y(\cdot;\vx)$ for the labels from the training sample $\calS$. We consider a fixed set of hypotheses $\calH$ for the rule $y(\cdot;\vx)$, and then select a hypothesis $h\in\calH$ such that it gives rise to the smallest generalization risk, 
\begin{align} \label{eRisk}
R(h) = \E_{\vx\sim P_X} \E_{\vlambda\sim P_\Lambda} \E_{y\sim P_{\bm{Y} | X\Lambda}} [h(\vlambda;\vx)\neq y] 
\end{align}
where $[\cdot]$ is the 0-1 loss. We provide a PAC bound in Section~\ref{sPAC} for the risk along with its empirical estimate from the training set $\calS$,
\begin{align} \label{eEmpRisk}
\widehat{R}(h) = \frac{1}{\cM\cL}\sum_{m=1}^\cM \sum_{l=1}^\cL L(h(\vlambda_l;\vx_m),y_{ml}) + \Omega(h),
\end{align}
where $L(\cdot)$ is a 0-1 loss (in practice one uses a surrogate differentiable loss), and $\Omega(h)$ is a regularization over $h$. At the high level, the risk of infinite-label learning can be understood as the traditional binary classification risk over the label $\vlambda$ and assignment $y|\vx$, conditioning on the variable $X$ representing the input data. 
}

\vspace{5pt}
\begin{theorem}
For any $\delta>0$, any $h(\x, \blambda) \in \H\triangleq\{\mathrm{sgn}\inner{V\x}{\blambda}~|~V\in \R^{\cst{n}\times \cst{d}}\}$, and under Assumptions I and  II, the following holds with probability at least $1-\delta$,
\begin{align}
\nonumber
&R(h) - \widehat{R}_\text{\em non-iid}(h) \\
 \leq &\mathcal{O}\Bigg(\sqrt{{\log \frac{1}{\delta} + \cst{d}\log\frac{\cM}{\cst{d}} \over \cM} + {\log \frac{\cM}{\delta} + \cst{n}\log \frac{\cL}{\cst{n}} \over \cL}}\Bigg).
\end{align}
\vspace{-15pt}
\end{theorem}
This generalization error bound is roughly $\mathcal{O}(1/\sqrt{\cM} + 1/\sqrt{\cL})$ if all $\log$ terms are ignored. When $\cM$ and $\cL$ converge to infinity, the error vanishes. An immediate implication is the learnability of the infinite-label learning: to obtain a certain accuracy on all labels --- both seen $\calL$ and unseen $\calU$, one does not have to observe all of them in the training phase. \eat{Another interesting implication for practice is that $\cM$ and $\cL$ should be chosen in the same order, because $\cM \gg \cL$ (or $\cM \ll \cL$) would not bring significant improvements to the generalization bound.} 

\vspace{-10pt}

\eat{
\footnote{Ji's comment: we may need to explain why this result is meaningful before REMARK. It is different from the traditional classification problem.}

Another note is that, with no surprise, the bound depends on both the number of data points and number of seen labels. In the next section we will empirically study the effect of varying the number of seen labels.

}

\begin{proof}
To prove the theorem, we essentially need to consider the following probability bound:
\begin{align}
& \P\Big(\exists~{h\in \H}: R(h)- \widehat{R}_\text{non-iid}(h) \geq \epsilon\Big) \nonumber
\\ 
\leq & \P\Big(\exists{h}: R(h)- A(\vlambda, y;\{\vx_m\}) \geq {\epsilon \over 2}\Big) \label{eq:proof:part1:0}
\\  
&+ \P\Big(\exists{h}:A(\vlambda, y;\{\vx_m\}) - \widehat{R}_\text{non-iid}(h) \geq {\epsilon \over 2}\Big) \label{eq:proof:part2:0}
\end{align}
where the last inequality is due to the fact $\P(a+b\geq \epsilon) \leq \P(a\geq \epsilon / 2) + \P(b\geq \epsilon / 2)$, and 
\begin{align}
\nonumber&A(\vlambda,y;\{\vx_m\}) \triangleq {1/ \cM}\sum_m  \E_{y,\vlambda|\x_m}[h(\x_m, \blambda) \neq y ]\\
\nonumber=& \,{1/ \cM}\sum_m  \E_{\sigma,\vlambda|\x_m}\big[h(\x_m, \blambda)\neq \sigma y^\star(\x_m,\blambda) \big] \\
\nonumber=& \,{1/ \cM}\sum_m  \E_{\sigma,\vlambda}\big[h(\x_m, \blambda)\neq \sigma y^\star(\x_m,\blambda) \big]\\
\nonumber=& \,\E_{\sigma,\vlambda}\Big({1/ \cM}\sum_m \big[h(\x_m, \blambda)\neq \sigma y^\star(\x_m,\blambda) \big]\Big).
\end{align}
The first equality follows {Assumption II} and the second is due to {Assumption I} as well as the independence between the binary random variable $\sigma$ and the data $\x_m$. 

Next we consider the bounds for the two terms \eqref{eq:proof:part1:0} and \eqref{eq:proof:part2:0}, respectively. For the first term \eqref{eq:proof:part1:0}, we have the following,
\begin{align}
\eqref{eq:proof:part1:0}
\nonumber=&\P\Big(\exists{h}: R(h)- A(\vlambda, y;\{\vx_m\}) \geq {\epsilon_1 }\Big) \\
=&\P\Big(\exists h: \E_{\sigma,\blambda} \E_{\x}\big[h(\x, \blambda) \neq \sigma y^\star(\vx,\vlambda)\big] \nonumber \\ 
\nonumber&- \E_{\sigma,\blambda}\Big({1 / \cM}\sum_m \big[h(\x_m, \blambda) \neq \sigma y^\star(\x_m, \blambda)\big]\Big) \geq {\epsilon_1}\Big)
\\ \nonumber 
\leq & 
\P\Big(\exists h, \exists \blambda, \exists \sigma:  \E_{\x}\big[h(\x, \blambda) \neq \sigma y^\star(\x, \blambda)\big] \\ 
&- {1 / \cM}\sum_{m} \big[h(\x_m, \blambda) \neq \sigma y^\star(\x_m, \blambda)\big] \geq {\epsilon_1}\Big).
\label{eq:proof:part1:1}
\end{align}

Denote $\big[h(\x_m, \blambda)\neq \sigma y^\star(\x_m, \blambda)\big]$ by a function $f(\x_m)$ given $\blambda$ and $\sigma$. We have $\E(f(\x)) = \E_{\x} [h(\x, \blambda) \neq y]$. Moreover, $f(\x_1), \cdots, f(\x_m)$ are i.i.d.\ random variables in $\{0,1\}$ thanks to the sampling procedure described in Section~\ref{sNonIID}. Define $\calF$ to be a hypothesis space 
\[
\calF\triangleq \Big\{f(\x) = \big[h(\x, \blambda)\neq \sigma y^\star(\vx,\vlambda)\big]~|~\forall h, \blambda, \sigma\Big\}.
\]
Using the new notations, we can rewrite \eqref{eq:proof:part1:1} by
\begin{align}
\label{eq:proof:part1}
&\nonumber\P\Big(\exists f \in \calF:~\E(f(\x)) - {1 / \cM}\sum_m f(\x_m) \geq {\epsilon_1}\Big)
\\  \leq &\;
4r(\calF, 2\cM) \times \exp(-\cM\epsilon_1^2/8) 
\end{align}
where the inequality uses the growth function bound~\citep[Section 3.2]{mohri2012foundations}, and $r(\calF, 2\cM)$ is the number of maximally possible configurations (or values of $\{f(\x_1), \cdots, f(\x_{2\cM})\}$) given $2\cM$ data points in $\R^\cst{d}$ over the hypothesis space $\calF$. Since $h(\x, \blambda)$ is in the form of $\text{sgn}\langle V\x, \blambda \rangle$, $r(\calF, 2\cM)$ is bounded by $r(\W, 2\cM)$ with $\W\triangleq\{\text{sgn}(\w^\top \x)~|~\w\in\R^\cst{d}\}$. Hence, we have
\begin{align}
\nonumber\text{\eqref{eq:proof:part1}} &\leq 4 r(\W, 2\cM) \times \exp(-\cM\epsilon_1^2/8) \\
& \leq 4\times ({2e\cM / \cst{d}})^\cst{d} \times \exp(-\cM\epsilon_1^2/8). 
\end{align}

Next, we find the bound for term \eqref{eq:proof:part2:0} which depends on {Assumption I} only.
\begin{align}
\eqref{eq:proof:part2:0}
\nonumber=&\P\Big(\exists{h}: A(\vlambda, y;\{\vx_m\}) -\widehat{R}_\text{non-iid}(h)\geq {\epsilon_2 }\Big) \\
\nonumber=& \P\Big(\exists h: {1/ \cM} \sum_m \E_{\blambda,y|\x_m} [h(\x_m, \blambda) \neq y] \\ 
\nonumber&\quad- {1/ \cM}\sum_m {1/ \cL} \sum_l [h(\x_m, \blambda_l) \neq y_{m,l}] \geq {\epsilon_2}\Big)\\
\nonumber \leq& \P\Big(\exists h,\exists m: \E_{\blambda,y|\x_m} [h(\x_m, \blambda) \neq y] \\ 
\nonumber&- {1/ \cL}\sum_l [h(\x_m, \blambda_l) \neq y_{m,l}] \geq {\epsilon_2}\Big) 
\\ \nonumber\leq &
\cM\times\P\Big(\exists h:~\E_{\blambda,y|\x_1} [h(\x_1, \blambda) \neq y] \\
&- {1/ \cL}\sum_l [h(\x_1, \blambda_l) \neq y_{1,l}] \geq {\epsilon_2}\Big). 
\label{eq:proof:part2:2}
\end{align}
where the last inequality uses the union bound. Denote $[h(\x_1, \blambda)\neq y]$ by $g(\blambda, y)$ for short. Then we have $\E(g(\blambda, y)) = \E_{ \blambda,y|\x_1} [h(\x_1, \blambda) \neq y]$ and $g(\blambda_1,y_{1,1}), \cdots, g(\blambda_L, y_{1,L})$ are i.i.d.\ variables taking values from $\{0,1\}$ given $\x_1$. Define $\G$ to be a hypothesis space
\begin{align*}
\G\triangleq{}\Big\{g(\blambda, y) = [h(\blambda, \x_1) \neq y]~|~h\in \H \Big\}.
\end{align*}
Then we can cast the probabilistic factor in \eqref{eq:proof:part2:2} into
\begin{align}
&\nonumber \P\Big(\exists g\in \G:~\E(g(\blambda, y)) - {1/ \cL}\sum_l g(\blambda_l, y_{l}) \geq {\epsilon_2}\Big)\\
\leq &
4 r(\G, 2\cL) \times \exp(-\cL\epsilon_2^2/8)
\end{align}
where the last inequality uses the growth function bound again. We omit the remaining parts of the proof; the supplementary material includes the complete proof.

\eat{
To bound $r(\G, 2\cL, \L)$, we consider the structure of $h(\x, \blambda) = \text{sgn}(V\x, \blambda)$ and define $\W' =\{\text{sgn}(\z^\top \blambda)~|~\z\in \R^n\}$. It suggests that 
\[
r(\G, 2\cL, \L) \leq r(\W', 2\cL, \R^n) \leq \Big({2e\cL \over n}\Big)^{\bold{VC}(\R^n)} = \Big({2e\cL \over n}\Big)^n.
\]
Hence we have
\[
\eqref{eq:proof:part2} \leq 4\cM\Big({2e\cL \over n}\Big)^n\times \exp(-\cL\epsilon_2^2/8).
\]
By letting the right hand side equal $\delta/2$, we have
\begin{align}
\label{eq:proof:part2end}
\epsilon_2 = \sqrt{8\log {8\cM/\delta} + 8 n\log (2e\cL/n) \over \cL}.
\end{align}
Plugging \eqref{eq:proof:part1end} and \eqref{eq:proof:part2end} into \eqref{eq:proof:part1:0} and \eqref{eq:proof:part2:0} respectively, we prove the theorem. 
}

\vspace{-10pt}

\end{proof}

%% file: exp.tex

\eat{
\paragraph{Thrust II: Algorithmic solutions.} For the second thrust, we plan to develop algorithmic solutions to infinite-label learning and strive to offer both  strong analytical properties and compelling practical performance. We first give a few examples of the hypothesis $h(\vlambda;\vx)$, demonstrating that $h(\vlambda;\vx)$ naturally deals with the zero-shot labels, and then present an algorithm to explicitly account for the incomplete annotations $\vy_m$ by employing group lasso.

Equipped with any of the above hypothesis set $\calH$, we are ready to show how to deal with the incomplete annotations by group lasso. Recall that the annotations $\vy_m$ of the training data $\vx_m$ reveal only part of the relevant labels in $\calL$. This notion can be captured by the constraints with slack variables $\{\xi_{m,lk}\}$ in the following optimization problem. 
\begin{align}
\min_{h\in\calH,\, \{\xi_{m,lk}\}}\quad & \sum_{m=1}^\cM \; \sum_{k:\,y_{mk}=-1} \left({\sum_{l:\,y_{ml}=+1} \xi_{m,lk}^2}\right)^{{1}/{2}} + \Omega(h), \\
\text{s.t.}  \quad & h(\lambda_l;\vx_m)\ge 1+h(\lambda_k;\vx_m)-\xi_{m,lk},\quad \forall m, \forall (l,k):\, y_{ml}=1 \,\&\, y_{mk}=-1 \\
\quad & \xi_{m,lk}\ge 0 \;\qquad\qquad\qquad\qquad \qquad\qquad \forall m, \forall (l,k):\, y_{ml}=1 \,\&\, y_{mk}=-1
\end{align}
Note that we group the slack variables for each irrelevant label $(y_{mk}=-1)$. In the solution, they tend to be all zeros within the group if the corresponding label is indeed irrelevant to the data point. Meanwhile, group lasso allows non-zero solutions for a few groups if they correspond to the actually relevant labels, avoiding overly penalizing them even though $y_{mk}=-1$ at training.
}

\begin{figure*}[t]
\includegraphics[width=0.34\textwidth]{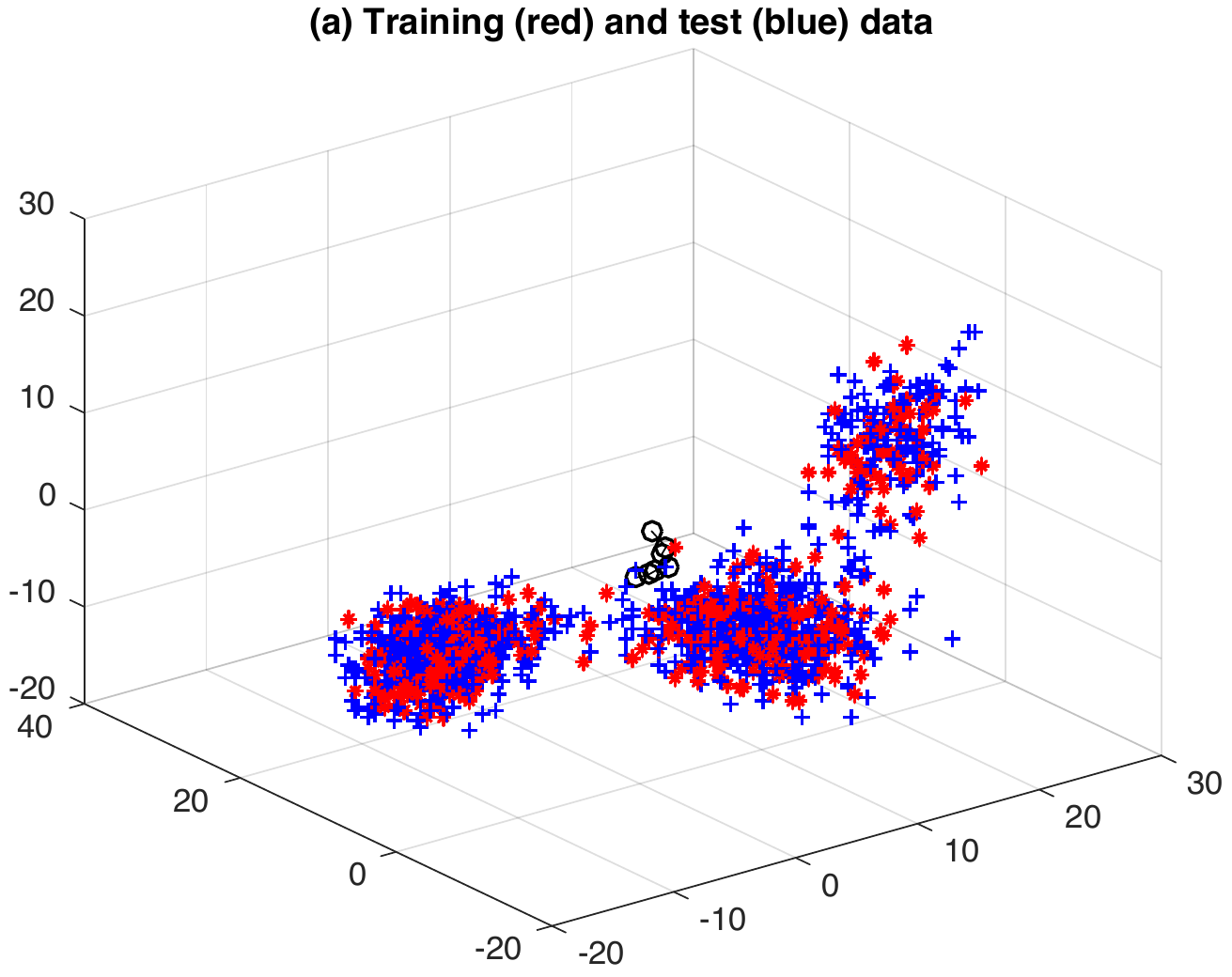}~~\includegraphics[width=0.3\textwidth]{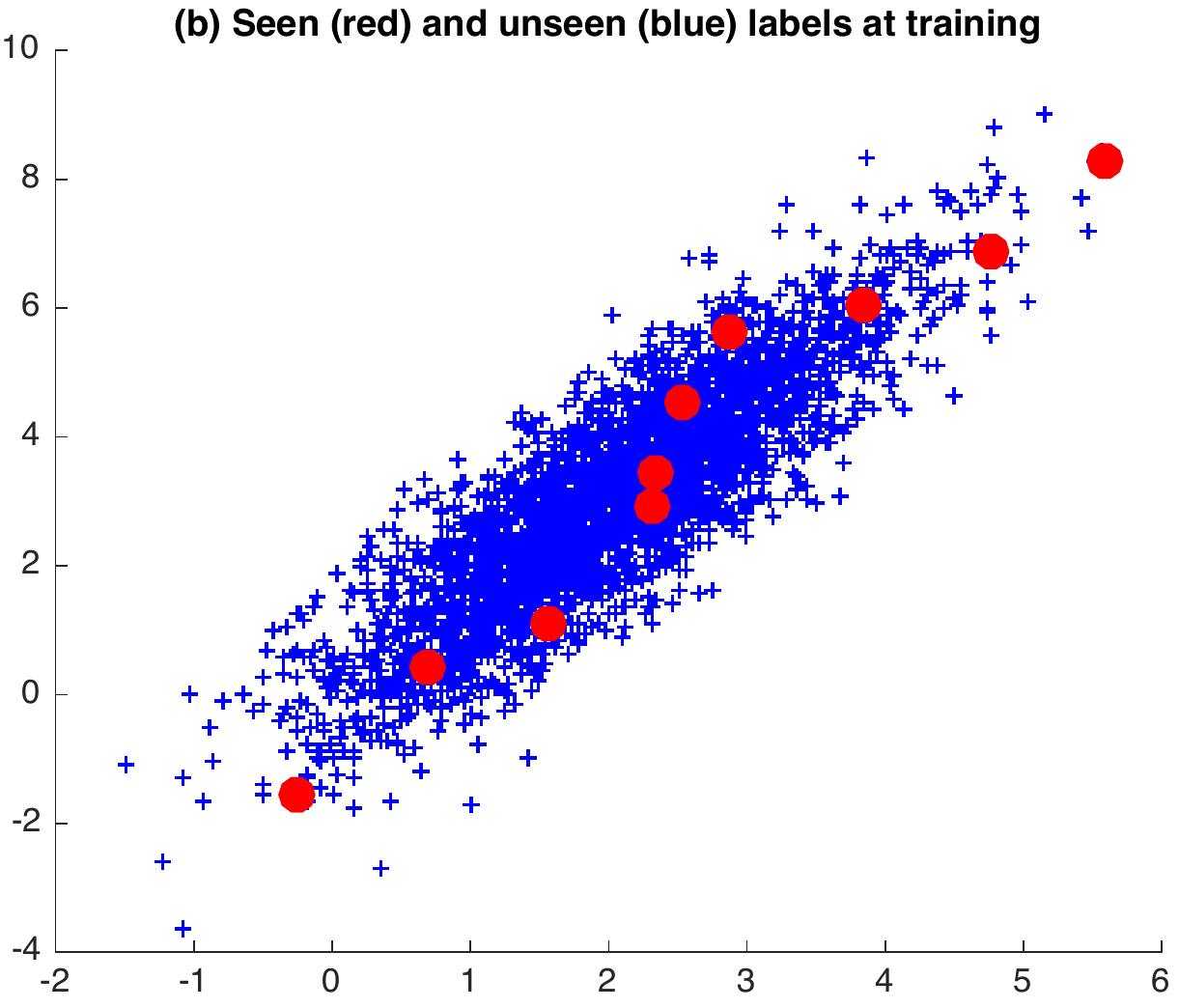}~~~\includegraphics[width=0.33\textwidth]{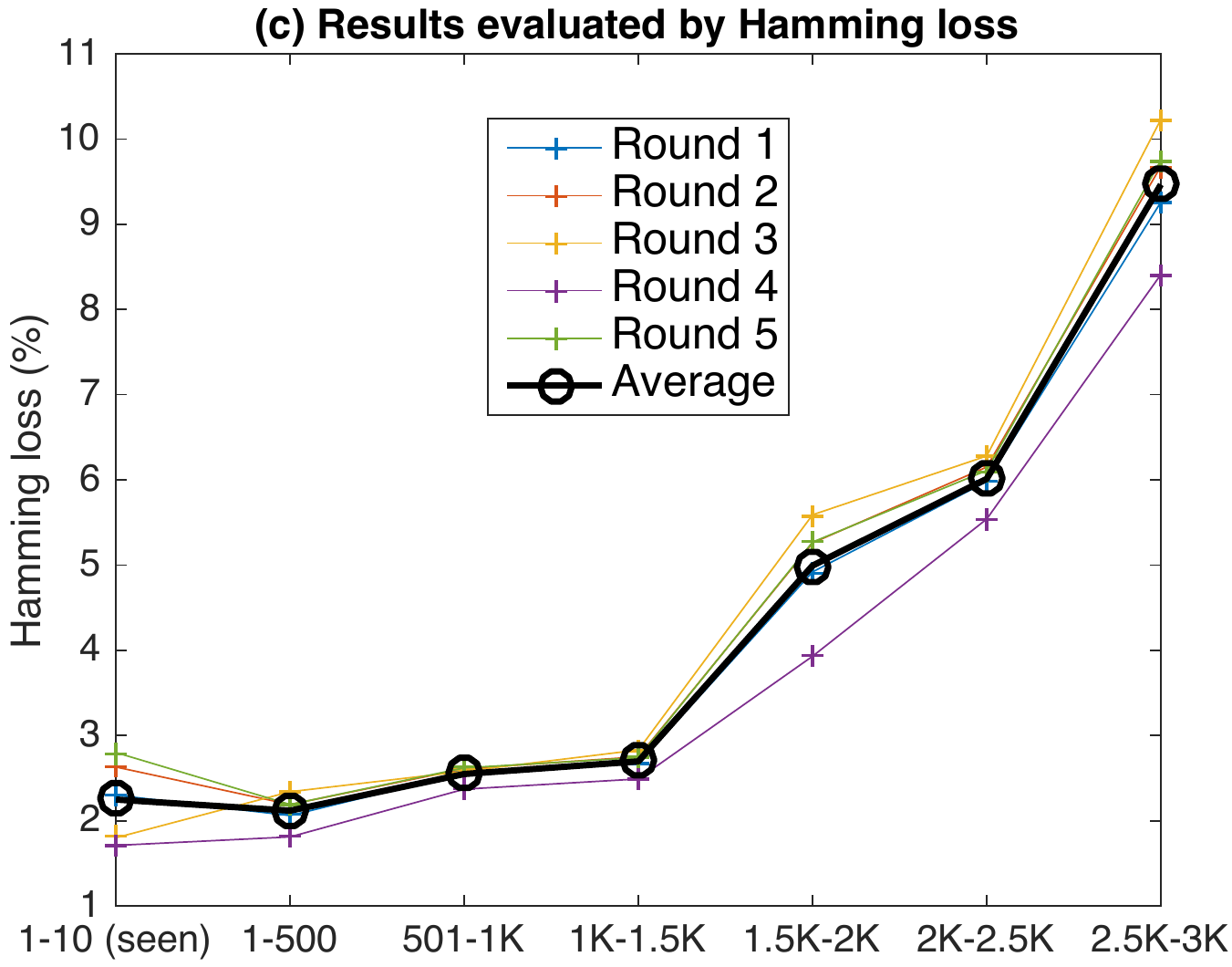}
\caption{Synthetic data points (a), labels (b), and experimental results given different unseen labels (c). The data and labels used in training are in the red color. We evaluate the results of labeling the test data with the unseen labels by the Hamming loss. The performance decreases as the number of  (Best viewed in color)} \label{fSynthetic}
\end{figure*}

\vspace{-10pt}
\section{Empirical studies} \label{sEmpirical} 
In this section, we continue to investigate the properties of infinite-label learning. While the theoretical result in Section~\ref{sPAC} justifies its learnability, there are many other questions of interest to explore for the practical use of the infinite-label learning. We focus on the following two questions and provide some empirical insights using respectively synthetic data and real data. 
\vspace{-5pt}
\begin{enumerate}  \setlength\itemsep{0.1em}
    \item After we learn a decision function from the training set, how many and what types of unseen labels can we confidently handle using this decision function?
    \item What is the effect of varying the number of seen labels $\calL$, given a fixed union $\calL\cup\calU$ of seen and unseen labels? Namely, given a labeling task, can we collect training data for only a subset of the labels and yet achieve good performance at the test phase on all the labels? We learn different decision functions by varying the number of seen labels $\calL$ and then check their labeling results of assigning all the labels $\calL\cup\calU$.  
\end{enumerate}

We use the bilinear form $h(\vx,\vlambda)=\text{sgn}\inner{V\vx}{\vlambda}$ for the decision function in the experiments, where our goal is not to achieve the state-of-the-art performance on the datasets but rather to demonstrate the feasibility of solving the infinite-label learning problems as well as revealing some trends and insights when the ratio between the numbers of seen and unseen labels changes. 

\eat{
\footnote{\jltext{The dimension for $\vx$ and $\vlambda$ are 3 and 2. Reviewers may be unhappy with that the dimension is too low. In addition, for the generalization error analysis, it would be more reasonable if $d/M$ and $n/L$ are in the same range. In our case, $d/M = 3/500$ and $n/L=2/10$.}}

\footnote{\jltext{What is the main purpose of the synthetic experiments? Is it for validating the idea ``learning the infinite-label'' from a limit number of observed labels? Does there exist any other algorithm? If so, we may need to compare with them. If no, we may create more figures to show that this idea works.}}
}

\subsection{Synthetic experiments}
To answer the first question, we generate some synthetic data which allows us to conveniently control the number of labels for the experiments.


\paragraph{Data.} We randomly sample 500 training data points $\{\vx_m\in\R^3\}_{m=1}^{\cM=500}$ and 1000 testing data points from a five-component Gaussian mixture model. We also sample 10 seen labels $\calL=\{\vlambda_l\in\R^2\}_{l=1}^{\cL=10}$ and additionally 2990 unseen labels $\calU=\{\vlambda_{11},\cdots,\vlambda_{3000}\}$ from a Gaussian distribution. Note that only the seen labels $\calL$ are revealed during the training stage. As below specifies the distributions,
\begin{align*}
\vx &\sim P(\vx) = \sum_{k=1}^5 \pi_k\, \calN\left(\vmu_k,\mU_k\mU_k^T\right), \quad \vx\in\R^3\\
\vlambda &\sim \calN\left(\left[\begin{array}{c}2\\3\end{array}\right],\left[\begin{array}{cc}1&1.5\\1.5&3\end{array}\right]\right), \quad \vlambda\in\R^2
\end{align*}
where the mixture weights $\pi_k$ are sampled from a Dirichlet distribution $\textsc{Dir}(3,3,3,3,3)$, and both the mean $\vmu_k$ and $\mU_k$ for the variance are sampled from a standard normal distribution (using the \texttt{randn} function in MATLAB). Finally, we generate a ``groundtruth'' matrix $V\in\R^{2\times 3}$ from a standard normal distribution. The groundtruth label assignments are thus given by $y^\star_{ml}=\text{sgn}\inner{V\vx_m}{\vlambda_l}$ for both training and testing data and both seen and unseen labels. Following \textbf{Assumption II}, we randomly flip the sign of each $y^\star_{ml}$ with probability $p=0.1$.

Figure~\ref{fSynthetic}(a) and (b) show the data points and labels we have sampled. The training data and the seen labels are in red color, while all the other (test) data points and labels are unseen during training. We choose the low dimensions for the data and vectorial labels so that we can visualize them and have a concrete understanding about the results to be produced.

\begin{table*}
\centering
\caption{Comparison of different methods on the image tagging task when the number of seen tags varies.}  \label{tRes}
\vspace{3pt}
\begin{tabular}{r|cccc|cccc|cccc}
\multirow{2}{*}{ Method~~~~\%}
 &\multicolumn{4}{c|}{\textit{81} out of 81 tags are seen} &\multicolumn{4}{c|}{\textit{73} out of 81 tags are seen} & \multicolumn{4}{c}{\textit{65} out of 81 tags are seen}\\
& MiAP & P & R  & { F1}& MiAP & P & R  & { F1}& MiAP & P & R  & { F1}  \\
    
\hline
 LabelEM~\cite{tsochantaridis2005large} & 47.4 & 26.2 & 44.7 & 33.1 & 41.8 & 23.4 & 39.8 & 29.4 & 38.4 & 21.4 & 36.4 & 26.9   \\

 ConSE~\cite{norouzi2013zero} & 47.5 & 26.5 & 44.9 & 33.2 & 46.9 & 26 & 44.3 & 32.7 & 44.9 & 24.3 & 41.5 & 30.7 \\

 ESZSL~\cite{romera2015embarrassingly} & 45.8 & 25.9 & 44.2 & 30.7 & 45.6 & 25.6 & 43.6 & 28.1 & 43.8 & 23.8 & 40.6 & 30.1 \\

{ Bilinear-RankNet} & \textbf{53.8} & \textbf{30.1} & \textbf{51.4} & \textbf{38} & \textbf{52.8} & \textbf{29.5} & \textbf{50.2} & \textbf{37.1} & \textbf{49.5} & \textbf{27.5} & \textbf{46.8} & \textbf{34.6}  \\
\hline
\end{tabular}
\end{table*}

\begin{table*}
\centering
\caption{Continuation of Table~\ref{tRes}, i.e., further comparisons of different methods on the image tagging task when the number of seen tags varies.} \label{tRes2}
\vspace{3pt}
\begin{tabular}{r|cccc|cccc|cccc}
\multirow{2}{*}{ Method~~~~\%}
 & \multicolumn{4}{c|}{\textit{57} out of 81 tags are seen}   &  \multicolumn{4}{c|}{\textit{49} out of 81 tags are seen}   & \multicolumn{4}{c}{\textit{41} out of 81 tags are seen}\\
    & MiAP & P & R  & { F1}& MiAP & P & R  & { F1}& MiAP & P & R  & { F1}  \\
\hline    

 LabelEM~\cite{tsochantaridis2005large} & 32.1 & 16.5 & 28.2 & 20.8 & 30.0 & 15.7 & 26.8 & 19.5 & 32.4 & 18.1 & 30.8 & 22.8   \\

 ConSE~\cite{norouzi2013zero} & 41.8 & 22.9 & 39.0 & 28.9 & 40.1 & 22.0 & 37.5 & 27.8 & 38.7 & 22.2 & 37.9 & 28.0   \\

 ESZSL~\cite{romera2015embarrassingly} & 41.6 & 22.8 & 38.9 & 28.7 & 39.6 & 21.7 & 36.9 & 27.3 & 38.4 & 21.9 & 37.4 & 27.6 \\

{ Bilinear-RankNet} & \textbf{46.8} & \textbf{26.3} & \textbf{44.8} & \textbf{33.1} & \textbf{45.1} & \textbf{25.8} & \textbf{44.0} & \textbf{32.5} & \textbf{41.2} & \textbf{23.7} & \textbf{40.3} & \textbf{29.8} \\
\hline
\end{tabular}
\end{table*}

\paragraph{Algorithm.} Given the training set $\calS$ of the 10 seen labels $\calL$, we learn the labeling function $h(\vx,\vlambda)=\text{sgn}\inner{V\vx}{\vlambda}$ by minimizing a hinge loss,
\[
\widehat{V} \leftarrow \arg\min_V \frac{1}{\cM\cL}\sum_{m=1}^{\cM=500} \sum_{l=1}^{\cL=10}\max\left(1-y_{ml}\inner{V\vx_m}{\vlambda_l}, 0\right),
\]
and then try to assign both seen and unseen labels $\calL\cup\calS$ to the $1000$ test data points, using $\widehat{V}$. It is interesting to note that a similar formulation and its dual have been studied by Szedmak et al.~\cite{szedmak2006learning}, which, however, are mainly used to  reduce the complexity of multi-class classification. It it also worth pointing out that it is not a regression problem though its form shares some similarity with the (multi-variate) support vector regression~\cite{drucker1997support}.

We incrementally challenge the learned infinite-label model $y_{ml} \leftarrow\text{sgn}\inner{\widehat{V}\vx_m}{\vlambda_l}$ by gradually increasing the difficulties at the test phase. Namely, we rank all the labels according to their distances to the seen labels $\calL$, where the distance between an unseen label $\vlambda_l$ and the seen ones $\calL$ is calculated by $\min_{\vlambda\in\calL}\|\vlambda-\vlambda_l\|_2$. We then evaluate the label assignment results for every 500 consecutive labels in the ranking list (as well as the 10 seen labels). Arguably, the last 500 labels, which are the furthest subset of unseen labels from the seen ones $\calL$, impose the biggest challenge to the learned model.  

\paragraph{Results.} Figure~\ref{fSynthetic}(c) shows the label assignment errors for different subsets of the test labels. We run 5 rounds of experiments each with different randomly sampled data, and report their individual results as well as the average. We borrow from multi-label classification~\cite{zhang2014review} the Hamming loss as the evaluation metric. It is computed by $\frac{1}{1000}\sum_{m=1}^{1000}(|\vy_m\neq\vy^\star_m|^T\bm{1})/|\vy^\star_m|^T\bm{1}$, \eat{ \footnote{\jltext{Is this something like precision? How about using AUC?}}} where $\vy^\star_m$ is the groundtruth label assignment to the data point $\vx_m$ and $\vy_m$ is the predicted assignment. Note that this is inherently different from the classification accuracy/error used for evaluating multi-class classification.  

We draw the following observations from Figure~\ref{fSynthetic}(c). First, infinite-label learning is feasible since we have obtained decent  results for up to 3000 labels with only 10 of them seen in training. Second, when the unseen labels are not far from the seen labels, the label assignment results are on par with the performance of assigning only the seen labels to the test data (cf.\ the Hamming losses over the first, second, and third 500 unseen labels). Third, labels that are far from the seen labels may cause larger confusions to the infinite-label  model learned from finite seen labels. Increasing the number of seen labels and/or data points during training can improve the model's generalization capability to unseen labels, as suggested by  Theorem 1 and revealed in the next experiment.

\begin{figure*}[t]
\centering
    \includegraphics[width=0.85\textwidth]{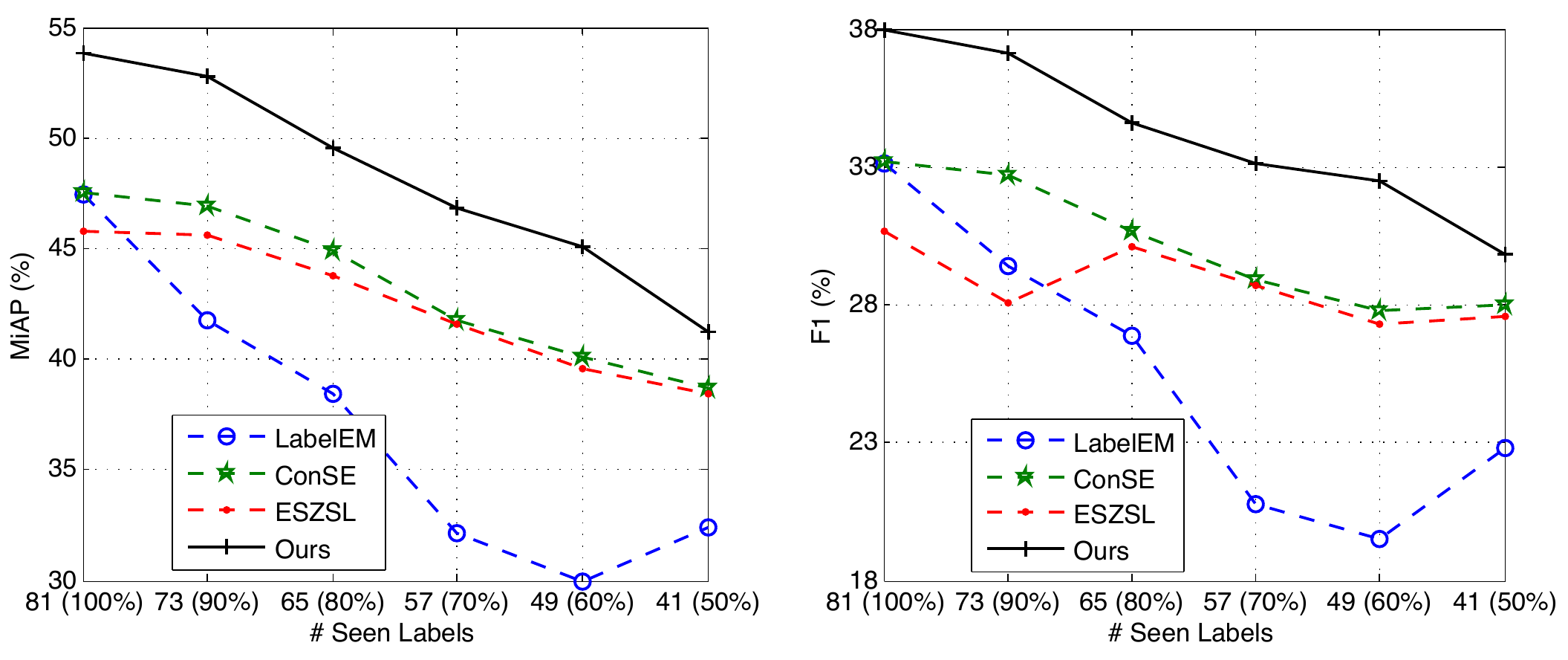}
    \vspace{-10pt}
  \caption{The infinite-label learning results for image tagging under different numbers of seen labels. } \label{fTag}
\end{figure*}

\subsection{Image tagging}
We experiment with image tagging to seek some clues to answer the second question raised at the beginning of this section. Suppose we are given a limited budget and have to build a tagging model as good as possible under this budget. Thanks to the infinite-label learning, we may compile a training set using only a subset of the labels of interest as opposed to asking the annotators to examine all of them. Then, how many seen labels should we use in order to achieve about the same performance as using all of the labels for training? We give some empirical results to answer this question. Our experiment protocol largely follows the prior work~\cite{Zhang_2016_CVPR}.

\paragraph{Data.} We conduct the experiments using the NUS-WIDE dataset~\cite{chua2009nus}. It has 269,648 images in total and we were only able to retrieve 223,821 of them using the provided image URLs. Among them, 134,281 are training images and the rest are testing images according to the official split by NUS-WIDE. We further randomly leave out 20\% of the training images as the validation set for tuning hyper-parameters. The images are each represented by the are the $l_2$-normalized activations of the last fully connected layer of VGGNet-19~\cite{simonyan2014very}. 

Each image in NUS-WIDE has been manually annotated with the relevant tags out of 81 candidate tags in total. We obtain the tags' vectorial representations using the pre-trained GloVe word vectors~\cite{pennington2014glove}. While all the 81 tags are considered at the test stage, we randomly choose $\cL=81$ (100\% out of the 81 tags), 73 (90\% out of the 81 tags), 65 (80\% out of 81), 57 (70\% out of 81), 49 (60\% out of 81), and 41 (50\% out of 81) seen tags for training different labeling functions $h(\vlambda;\vx)=\text{sgn}\inner{V\vx}{\vlambda}$. If a training image has no relevant tags under some of the settings, we simply drop it out of the training set.

\paragraph{Learning and evaluation.}  Image tagging is often evaluated based on the top few tags returned by a system, assuming that users do not care about the remaining ones. We report the results measured by four popular metrics: Mean image Average Precision (MiAP)~\cite{li2015socializing} and the top-3 precision, recall, and F1-score. Accordingly, in order to impose the ranking property to our labeling function, we learn it using the RankNet loss~\cite{burges2005learning},
\begin{align}
\nonumber &\widehat{V} \leftarrow \arg\min_V \frac{1}{\cM} \sum_{m=1}^{\cM} \frac{1}{(\cL-{\cK_m})\cK_m}  \sum_{\substack{k:y_{mk}=+1 \\
\bar{k}:y_{m\bar{k}}=-1}
} 
\\ \nonumber&\log(1+\exp(\inner{V\vx_m}{\vlambda_{\bar{k}}}-\inner{V\vx_m}{\vlambda_k}))+\gamma\norm{V}_2,
\end{align}
where $\cK_m$ is the number of relevant tags to the $m$-th image. The hyper-parameter $\gamma$ is tuned by the MiAP results on the validation set.

\eat{
 We performed two tagging experiments on four subsets of NUS-WIDE dataset~\cite{chua2009nus} dataset.
 The first experiment is zero-shot learning tagging task, in which a model is learned on seen tags and used to predicted a disjoint unseen tags set. In order to organize zero-shot tagging scenarios, we then create four subsets by subsampling the dataset though separating $16/24/32/40$ labels from the 81 labels as unseen labels we used for evaluation on testing images. And the rest $65/57/49/41$ labels would be the $\cL$ seen labels we used to train the model with training images. Images that is not labeled after subsampling are drop from training/validation/evaluation.
 Th second experiment is seen/unseen tagging task. Models in this task are same in the first experiment. However instead of tested on disjoint unseen tags set, model are evaluated on the whole 81 labels, which is the mixture of both seen and unseen tags. This senarios is closer to the reality in which both seen and unseen tags are expected to appear during testing.
}

\paragraph{Baselines.}
We compare our results to those of three state-of-the-art zero-shot learning methods: LabelEM~\cite{akata2015evaluation} whose compatible function is the same as ours except that it learns the model parameters by structured SVM~\cite{tsochantaridis2005large}, ConSE~\cite{norouzi2013zero} which estimates the representations of the unseen labels using the weighted combination of the seen ones', and ESZSL~\cite{romera2015embarrassingly} that enjoys a closed form solution thanks to the choice of the Frobenius norm. 

The methods above were developed for multi-class classification, namely, to exclusively classify an image to one and only one of the classes. As a result, when we train them using the image tagging data, different tags may have duplicated images and result in conflict terms in the objective functions. We resolve this issue by removing such conflict terms in training. By doing this, we observe about 0.5\%--2\% absolute improvements for the baselines over blindly applying them to our infinite-label learning problem.

\paragraph{Results.} 
Tables~\ref{tRes} and~\ref{tRes2} present the results of our method and the competitive baselines evaluated by the MiAP, top-3 precision, recall, and F1-score. We also plot the MiAPs and F1-scores in Figure~\ref{fTag} to visualize the differences between different methods and, more importantly, the changes over different numbers of seen labels. Recall that no matter how many labels are seen in the training set, the task remains the same at the test phase, i.e., to find all relevant labels from the 81 candidate tags for each image.


We can see that the performances of all the methods decrease as fewer seen tags are used for training. However, the performance drop is fairly gentle for our method---the MiAP drops by 1\% (respectively, 3\%) absolutely from using 100\% seen labels for training to using 90\% (respectively, from 90\% to 80\%). This confirms our conjecture that, with the semantic codes of the labels, we may save some human labor efforts for annotating the data without sacrificing the overall performance at the test phase.

Additionally, we find that our learned decision function outperforms all the competitive baseline methods by a large margin no matter under which experiment setting. This mainly attributes to that  it is learned by a tailored loss function for the infinite-label learning problem, while the others are not at all. To some extent, such results affirm the significance of studying the infinite-label learning as a standalone problem because the existing methods for zero-shot classification are suboptimal for this new  learning framework.


\eat{
The quantitative ranking results are presented in Table.~\ref{tZSL} and Table.~\ref{tMixedL}. Besides aforementioned baselines, an upper bound {\tt Traditional} is also presented. It is a conventional tagging linear models trained on the corresponding unseen tags, instead of seen tags, in each experiment with training images. Overall our posed work outperform other state-of-the-art methods.
}

\eat{
LabelEM+. ConSE is a decent zero-shot classification method. If we consider each image tag as a unique class, then we could rank the tags by introducing zero-shot classification method such as ConSE. LabelEM+ is an tagging variation of zero-shot classification method LabelEM~\cite{}. In this variation we remove the term involve duplicated images resulted from considering each tag as class.
}

\eat{
\begin{table*}
    \centering
\caption{Comparison of \textbf{Zero-Shot} image tagging tasks results}
\vspace{0pt}
\label{tZSL}
\small
\makebox[0pt]{
\begin{tabular}{|l|c|c|c|c||c|c|c|c||c|c|c|c||c|c|c|c|}
\hline
\multirow{4}{*}{ Method~~~~\%}& \multicolumn{16}{|c|}{\textbf{Training} tags / \textbf{Testing} tags}\\
 & \multicolumn{4}{|c||}{\textbf{65} Seen tags / \textbf{16} Unseen tags} & \multicolumn{4}{c||}{\textbf{57} Seen tags / \textbf{24} Unseen tags}   &  \multicolumn{4}{c||}{\textbf{49} Seen tags / \textbf{32} Unseen tags}   & \multicolumn{4}{c|}{\textbf{41} Seen tags / \textbf{40} Unseen tags}\\
\hhline{~----------------} 
& \multirow{2}{*}{MiAP} & \multicolumn{3}{c||}{Top 3}  & \multirow{2}{*}{ MiAP} & \multicolumn{3}{c||}{Top 3}  & \multirow{2}{*}{ MiAP} & \multicolumn{3}{c||}{Top 3}  & \multirow{2}{*}{ MiAP} & \multicolumn{3}{c|}{Top 3}\\
    & & P & R  & { F1} & & P & R & { F1} & & P & R & { F1} & & P & R & { F1} \\
    
\hline\hline
{\tt Traditional} & {\tt80.3} & {\tt30.9} & {\tt88.0} & {\tt45.7} & {\tt75.2} & {\tt30.5} & {\tt82.6} & {\tt44.6} & {\tt70.8} & {\tt28.6} & {\tt76.9} & {\tt41.7} & {\tt63.8} & {\tt28.8} & {\tt67.5} & {\tt36.1}  \\
\hline
 LabelEM+ & 28.9 & 11.2 & 31.9 & 16.6 & 37.0 & 14.5 & 39.2 & 21.2 & 25.9 & 10.3 & 27.7 & 15.0 & 24.7 & 11.1 & 25.9 & 15.5   \\
\hline
 ConSE & 63.6 & 24.0 & 68.4 & 35.5 & 50.6 & 21.0 & 56.8 & 30.6 & 46.7 & 18.2 & 48.9 & 26.6 & 40.2 & 17.7 & 41.5 & 24.8   \\
\hline

{ Proposed} & \textbf{64.6} & \textbf{25.0} & \textbf{71.1} & \textbf{37.0} & \textbf{54.0} & \textbf{22.7} & \textbf{61.5} & \textbf{33.2} & \textbf{52.1} & \textbf{21.3} & \textbf{57.3} & \textbf{31.1} & \textbf{42.7} & \textbf{19.4} & \textbf{45.5} & \textbf{27.2} \\
\hline
\end{tabular}
}
\end{table*}
}

\eat{
\begin{table*}
    \centering
\caption{Comparison of \textbf{Sseen/Unseen} image tagging tasks results}
\vspace{0pt}
\label{tMixedL}
\small
\makebox[0pt]{
\begin{tabular}{|l|c|c|c|c||c|c|c|c||c|c|c|c||c|c|c|c|}
\hline
\multirow{4}{*}{ Method~~~~\%}& \multicolumn{16}{|c|}{\textbf{Training} tags / \textbf{Testing} tags}\\
 & \multicolumn{4}{|c||}{\textbf{65} Seen tags / 81 Mixed tags} & \multicolumn{4}{c||}{\textbf{57} Seen tags / 81 Mixed tags}   &  \multicolumn{4}{c||}{\textbf{49} Seen tags / 81 Mixed tags}   & \multicolumn{4}{c|}{\textbf{41} Seen tags / 81 Mixed tags}\\
\hhline{~----------------} 
& \multirow{2}{*}{MiAP} & \multicolumn{3}{c||}{Top 3}  & \multirow{2}{*}{ MiAP} & \multicolumn{3}{c||}{Top 3}  & \multirow{2}{*}{ MiAP} & \multicolumn{3}{c||}{Top 3}  & \multirow{2}{*}{ MiAP} & \multicolumn{3}{c|}{Top 3}\\
    & & P & R  & { F1} & & P & R & { F1} & & P & R & { F1} & & P & R & { F1} \\
    
\hline\hline
 LabelEM+ & 38.4 & 21.4 & 36.4 & 26.9 & 32.1 & 16.5 & 28.2 & 20.8 & 30.0 & 15.7 & 26.8 & 19.5 & 32.4 & 18.1 & 30.8 & 22.8   \\
\hline
 ConSE & 44.9 & 24.3 & 41.5 & 30.7 & 41.8 & 22.9 & 39.0 & 28.9 & 40.1 & 22.0 & 37.5 & 27.8 & 38.7 & 22.2 & 37.9 & 28.0   \\
\hline

{ Proposed} & \textbf{49.5} & \textbf{27.5} & \textbf{46.8} & \textbf{34.6} & \textbf{46.8} & \textbf{26.3} & \textbf{44.8} & \textbf{33.1} & \textbf{45.1} & \textbf{25.8} & \textbf{44.0} & \textbf{32.5} & \textbf{41.2} & \textbf{23.7} & \textbf{40.3} & \textbf{29.8} \\
\hline
\end{tabular}
}
\end{table*}
}

%% file: conclusion.tex
\section{Conclusion} \label{sConclusion}
In this paper, we study a new machine learning framework, infinite-label learning. It fundamentally expands the scope of multi-label learning in the sense that  the learned decision function can assign both seen and unseen labels to a data point from potentially an infinite number of candidate labels. We provide a formal treatment to the infinite-label learning, discuss its distinction from existing machine learning settings, lay out mild assumptions to a PAC-learning bound for the new problem, and also empirically examine its feasibilities and properties.

There are many  potential avenues for the future work. Our current PAC bound can be likely improved and the assumptions could be relaxed. Theoretical understanding about its performance under the MiAP evaluation is also necessary given that MiAP is prevalent in evaluating the multi-label learning results, especially for image tagging. One particularly interesting application of the infinite-label learning is on the extreme multi-label classification problems~\cite{bhatia2015sparse}. We will explore them in the future work.

%% file: theorem-supp.tex

We examine the theoretical properties of zero-shot multi-label learning under the PAC learning framework. Given the training sample $\calS=\{(\vx_m,\vy_m)\in\R^d\times\{-1,+1\}^\cL\}_{m=1}^\cM$ and the semantic codes of the seen labels $\calL=\{\vlambda_l\in\R^n\}_{l=1}^\cL$, the theorem below sheds lights on the numbers of data points and labels that are necessary to achieve a particular level of error. 

\begin{theorem}
For any $\delta>0$, any $h(\x, \blambda) \in \H:=\{\mathrm{sgn}\inner{V\x}{\blambda}~|~V\in \R^{n\times d}\}$, and under Assumptions I and II presented in the main text, the following holds with probability at least $1-\delta$,
\begin{align}
\nonumber
&\E_\x\E_\blambda\E_{y|\x,\blambda} [h(\x, \blambda) \neq y] - {1\over \cM\cL} \sum_{m=1}^\cM\sum_{l=1}^\cL [h(\x_m, \blambda_l) \neq y_{m,l}] \\
 &\leq 2\,\max\Bigg(\sqrt{8\log (8/\delta) + 8d\log(2e\cM/d) \over \cM},\Bigg. \\ & \quad\quad \Bigg.\sqrt{8\log ({8\cM/\delta}) + 8 n\log (2e\cL/n) \over \cL}\;\Bigg).
\end{align}
\end{theorem}

\newpage

\eat{
\footnote{Ji's comment: we may need to explain why this result is meaningful before REMARK. It is different from the traditional classification problem.}

Another note is that, with no surprise, the bound depends on both the number of data points and number of seen labels. In the next section we will empirically study the effect of varying the number of seen labels.

}

\begin{proof}
To prove the theorem, we essentially need to consider the following probability bound:
\begin{align}
\nonumber
& \P\bigg(\exists~{h\in \H}:~ \E_\x\E_\blambda\E_{y|\x, \blambda}[h(\x, \blambda) \neq y] \\ & \quad- {1\over \cM\cL} \sum_{m=1}^\cM\sum_{l=1}^\cL [h(\x_m, \blambda_l) \neq y_{m,l})] \geq \epsilon\bigg)
\\ \nonumber = & 
\P\Bigg(\exists~{h\in \H}:~\E_\x\E_\blambda\E_{y|\x,\blambda}[h(\x, \blambda) \neq y] \\ \nonumber &\quad - {1\over \cM}\sum_{m=1}^\cM\E_\blambda \E_{y|\x_m,\blambda}[h(\x_m, \blambda) \neq y] 
\\ \nonumber
& \quad+ {1\over \cM}\sum_{m=1}^\cM\E_\blambda \E_{y|\x,\blambda}[h(\x_m, \blambda) \neq y] \\ \nonumber &\quad- {1\over \cM\cL} \sum_{m=1}^\cM\sum_{l=1}^\cL [h(\x_m, \blambda_l) \neq y_{m,l}] \geq \epsilon\Bigg)
\\ \nonumber
\leq & \P\Bigg(\exists~{h\in \H}:~\E_\x\E_\blambda\E_{y|\x,\blambda}[h(\x, \blambda) \neq y] \\ &\quad- {1\over \cM} \sum_{m=1}^{\cM} \E_\blambda \E_{y|\x_m,\blambda}[h(\x_m, \blambda) \neq y ] \geq {\epsilon \over 2}\Bigg) \label{eq:proof:part1:0}
\\  \nonumber&
+ \P\Bigg(\exists~{h\in \H}:~{1\over \cM}\sum_{m=1}^{\cM} \E_\blambda \E_{y|\x_m,\blambda}[h(\x_m, \blambda) \neq y ] \\ &\quad- {1\over \cM\cL} \sum_{m,l} [h(\x_m, \blambda_l) \neq y_{m,l}] \geq {\epsilon \over 2}\Bigg) \label{eq:proof:part2:0}
\end{align}
where the last inequality is due to the fact $\P(a+b\geq \epsilon) \leq \P(a\geq \epsilon / 2) + \P(b\geq \epsilon / 2)$. Next we consider the bound for the two terms \eqref{eq:proof:part1:0} and \eqref{eq:proof:part2:0} above respectively. 

For the first term \eqref{eq:proof:part1:0}, we note that the random variable $y$ depends on the true label of $(\x, \blambda)$ and a random variable $\sigma$ in the way: $y = \sigma(\x, \blambda)$ where $y(\x, \blambda)$ is the true label of $(\x, \blambda)$ and $\sigma \in \{1,-1\}$ is a binary random variable independent to $\x$ and $\blambda$. Then we have
\begin{align}
\nonumber
&\P\bigg(\exists h \in \H:~\E_{\x,\blambda, y}[h(\x, \blambda) \neq y] \\ \nonumber&\quad- {1\over \cM}\sum_{m=1}^\cM \E_{\blambda, y | \x_m}[h(\x_m, \blambda) \neq y] \geq {\epsilon_1}\bigg)
\\ \nonumber
= &
\P\Bigg(\exists h \in \H:~\E_{\x,\blambda, \sigma}[h(\x, \blambda) \neq \sigma y(\x, \blambda)] \\ \nonumber&\quad- {1\over \cM}\sum_{m=1}^\cM \E_{\sigma, \blambda | \x_m}[h(\x_m, \blambda) \neq \sigma y(\x_m, \blambda)] \geq {\epsilon_1}\Bigg)
\\ \nonumber
= &
\P\Bigg(\exists h \in \H:~\E_{\x,\blambda, \sigma}[h(\x, \blambda) \neq \sigma y(\x, \blambda)] \\ \nonumber&\quad- {1\over \cM}\sum_{m=1}^\cM \E_{\sigma, \blambda}[h(\x_m, \blambda) \neq \sigma y(\x_m, \blambda)] \geq {\epsilon_1}\Bigg)
\\ \nonumber
= &
\P\Bigg(\exists h \in \H:~\E_{\sigma,\blambda} \E_{\x}[h(\x, \blambda) \neq y] \\ \nonumber&\quad- \E_{\sigma,\blambda}\Bigg({1\over \cM}\sum_{m=1}^\cM [h(\x_m, \blambda) \neq \sigma y(\x_m, \blambda)]\Bigg) \geq {\epsilon_1}\Bigg)
\\ \leq & 
\P\Bigg(\exists h \in \H, \exists \blambda \in \L, \exists \sigma \in \{-1, 1\}: \\ \nonumber&\quad\quad \E_{\x}[h(\x, \blambda) \neq \sigma y(\x, \blambda)] \\ &\quad- {1\over \cM}\sum_{m=1}^\cM [h(\x_m, \blambda) \neq \sigma y(\x_m, \blambda)] \geq {\epsilon_1}\Bigg)
\label{eq:proof:part1:1}
\end{align}
where the first equality is due to the dependence between $\blambda$ and $\x_m$.

Denote $[h(\x_m, \blambda)\neq \sigma y(\x_m, \blambda)]$ by a function $f(\x_m)$ given $\blambda$ and $\sigma$. We have $\E(f(\x_m)) = \E_{\x} [h(\x, \blambda) \neq y]$. $f(\x_1), \cdots, f(\x_m)$ are i.i.d. random variables in $\{0,1\}$. Define $\calF$ to be a hypothesis space 
\[
\calF:= \{f(\x) = [h(\x, \blambda)\neq y]~|~h \in \H,~\blambda \in \L\}.
\]
Using the new notation, we can rewrite \eqref{eq:proof:part1:1} in the following
\begin{align}
\label{eq:proof:part1}
&\nonumber\P\Bigg(\exists f \in \calF:~\E(f(\x)) - {1\over \cM}\sum_{m=1}^\cM f(\x_m) \geq {\epsilon_1}\Bigg)
\\  \leq &\;
4r(\calF, 2\cM, \X) \times \exp(-\cM\epsilon_1^2/8) 
\end{align}
where the inequality uses the Growth function bound, and $r(\calF, 2\cM, \X)$ is number of maximally possible configurations (or values of $\{f(\x_1), \cdots, f(\x_{2\cM})\}$) given $2\cM$ samples in $\X$ over the hypothesis space $\calF$. Since $h(\x, \blambda)$ is in the form of $\text{sgn}\langle V\x, \blambda \rangle$, $r(\calF, 2\cM, \X)$ is bounded by $r(\W, 2\cM, \X)$ with $\W:=\{\text{sgn}(\w^\top \x)~|~\w\in\R^d\}$. Hence, we have
\begin{align}
\nonumber r(\mathcal{F}, 2\cM, \X) \leq r(\W, 2\cM, \X) & \leq \Bigg({2e\cM\over d}\Bigg)^{{\bold{VC}}(\R^d)} \\&= \Bigg({2e\cM\over d}\Bigg)^{d}
\end{align}
where $\bold{VC}(\R^d)$ is the VC dimension for the space $\R^d$. We then have 
\begin{align*}
\text{\eqref{eq:proof:part1}} \leq 4 \Bigg({2e\cM \over d}\Bigg)^{d} \times \exp(-\cM\epsilon_1^2/8).
\end{align*}
By letting the right hand side equal  $\delta/2$, we have
\begin{align}
\label{eq:proof:part1end}
\epsilon_1 =  \sqrt{8\log {8/\delta} + 8d\log(2e\cM/d) \over \cM}.
\end{align}

Next we consider the upper bound for \eqref{eq:proof:part2:0}:
\begin{align}
\label{eq:proof:part2}
& \nonumber\P\Bigg(\exists h\in \H~:~{1\over \cM} \sum_{m=1}^\cM \E_{\blambda,y|\x_m} [h(\x_m, \blambda) \neq y] \\ &\quad- {1\over \cM}\sum_{m=1}^\cM {1\over \cL} \sum_{l=1}^\cL [h(\x_m, \blambda_l) \neq y_{m,l}] \geq {\epsilon_2}\Bigg)
\\ \nonumber = & 
\P\Bigg(\exists h\in \H~:~{1\over \cM} \sum_{m=1}^\cM \E_{\blambda} \E_{y|\x_m, \blambda} [h(\x_m, \blambda) \neq y] \\ \nonumber&\quad- {1\over \cM}\sum_{m=1}^\cM {1\over \cL} \sum_{l=1}^\cL [h(\x_m, \blambda_l) \neq y_{m,l}] \geq {\epsilon_2}\Bigg)
\\ \nonumber \leq &
\P\Bigg(\exists h\in \H,~\exists m\in \{1,\cdots, \cM\}:~\E_{\blambda}\E_{y|\x_m, \blambda} [h(\x_m, \blambda) \neq y] \\ \nonumber&\quad- {1\over \cL}\sum_{l=1}^\cL [h(\x_m, \blambda_l) \neq y_{m,l}] \geq {\epsilon_2}\Bigg) 
\\ \nonumber\leq &
\cM\times\P\Bigg(\exists h\in \H:~\E_{y, \blambda|\x_1} [h(\x_1, \blambda) \neq y] \\ &\quad- {1\over \cL}\sum_{l=1}^\cL [h(\x_1, \blambda_l) \neq y_{1,l}] \geq {\epsilon_2}\Bigg). 
\label{eq:proof:part2:2}
\end{align}
where the last inequality uses the union bound. Denote $[h(\x_1, \blambda)\neq y]$ by $g(\blambda, y)$ for short. Then we have $\E(g(\blambda, y)) = \E_{y, \blambda|\x_1} [h(\x_1, \blambda) \neq y]$ and $g(\blambda_1,y_{1,1}), \cdots, g(\blambda_L, y_{1,L})$ are i.i.d. samples taking values from $\{0,1\}$ given $\x_1$. Define $\G$ to be a hypothesis space
\begin{align*}
\G:=\{g(\blambda, y) = [h(\blambda, \x_1) \neq y]~|~h\in \H \}.
\end{align*}
Then we can cast the probabilistic factor in \eqref{eq:proof:part2:2} into
\begin{align}
&\nonumber \P\Bigg(\exists g\in \G:~\E(g(\blambda, y)) - {1\over \cL}\sum_{l=1}^\cL g(\blambda_l, y_{l}) \geq {\epsilon_2}\Bigg)\\
\leq &
4 r(\G, 2\cL, \L) \times \exp(-\cL\epsilon_2^2/8)
\end{align}
where the last inequality uses the Growth function bound again. To bound $r(\G, 2\cL, \L)$, we consider the structure of $h(\x, \blambda) = \text{sgn}(V\x, \blambda)$ and define $\W' =\{\text{sgn}(\z^\top \blambda)~|~\z\in \R^n\}$. It suggests that 
\[
r(\G, 2\cL, \L) \leq r(\W', 2\cL, \R^n) \leq \Bigg({2e\cL \over n}\Bigg)^{\bold{VC}(\R^n)} = \Bigg({2e\cL \over n}\Bigg)^n.
\]
Hence we have
\[
\eqref{eq:proof:part2} \leq 4\cM\Bigg({2e\cL \over n}\Bigg)^n\times \exp(-\cL\epsilon_2^2/8).
\]
By letting the right hand side equal $\delta/2$, we have
\begin{align}
\label{eq:proof:part2end}
\epsilon_2 = \sqrt{8\log {8\cM/\delta} + 8 n\log (2e\cL/n) \over \cL}.
\end{align}
Plugging \eqref{eq:proof:part1end} and \eqref{eq:proof:part2end} into \eqref{eq:proof:part1:0} and \eqref{eq:proof:part2:0} respectively, we prove the theorem.

\end{proof}